\documentclass[letterpaper, 10 pt, conference]{ieeeconf}  %

\IEEEoverridecommandlockouts                              %

\usepackage{times} %
\usepackage{amsmath} %
\usepackage{amssymb}  %

\usepackage{hyperref}
\usepackage{url}

\usepackage{epsfig}
\usepackage{graphicx}
\usepackage{amsmath}
\usepackage{amssymb}

\usepackage{wrapfig}

\usepackage{color}
\usepackage{xcolor}
\usepackage{capt-of}
\usepackage{caption}
\usepackage[ruled]{algorithm2e}
\usepackage{gensymb}
\usepackage{here}

\usepackage{cuted}
\usepackage{capt-of}

\newcommand{\ie}{\emph{i.e.}}
\newcommand{\eg}{\emph{e.g.}}
\newcommand{\etc}{etc}

\DeclareMathOperator*{\argmin}{arg\,min}

\title{\LARGE \bf
ViFu: Multiple 360$\degree$ Objects Reconstruction \\ with Clean Background via Visible Part Fusion}

\author{Tianhan Xu$^{1}$,%
\thanks{$^{1}$ The University of Tokyo, 7-3-1 Hongo, Bunkyo-ku, Tokyo, 113-8656, Japan. {\tt\small tianhan.xu@mi.t.u-tokyo.ac.jp}}
Takuya Ikeda$^{2}$, Koichi Nishiwaki$^{2}$
\thanks{$^{2}$ All authors are with the Woven by Toyota. 3 Chome-2-1 Nihonbashimuromachi, Chuo City, Tokyo 103-0022, Japan, {\tt\small [firstname.lastname]@woven.toyota}}
\thanks{This work has been submitted to the IEEE for possible publication. Copyright may be transferred without notice, after which this version may no longer be accessible.}
}

\begin{document}

\maketitle
\begin{strip}\centering
\vspace{-5em}
\includegraphics[width=\textwidth]{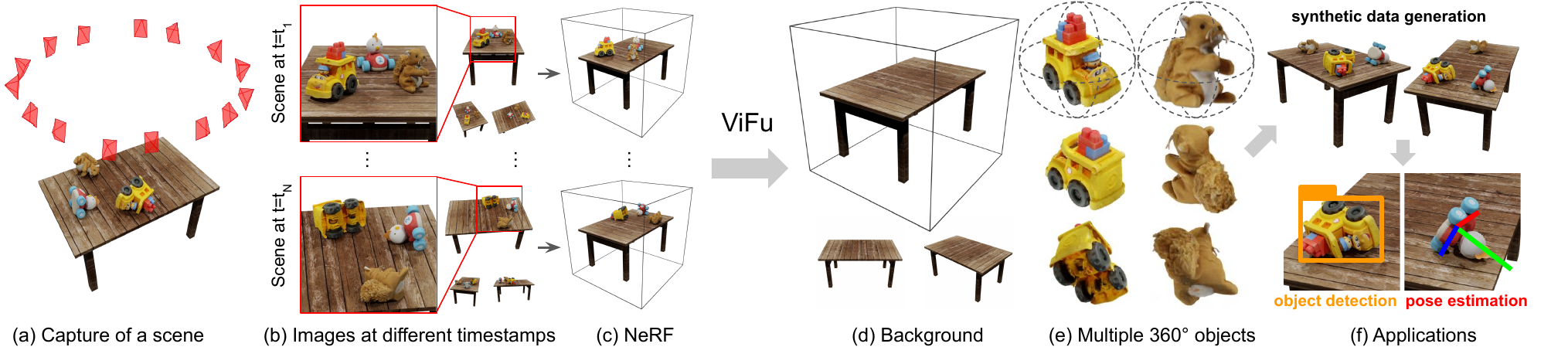}
\captionof{figure}{\textbf{An overview of our approach}. (a,b) By capturing multi-view images of the scenes at different timestamps, ViFu recovers the appearance and 3D geometry of (d) clean static backgrounds and (e) multiple 360$^{\circ}$ foreground objects. 
(c) NeRF representation supports free-view rendering for clean backgrounds and multiple foreground objects, including their rearrangement, thereby (f) facilitating the datasets creation for downstream tasks.
\label{fig:teaser}}
\vspace{-1em}
\end{strip}

\begin{abstract}
In this paper, we propose a method to segment and recover a static, clean background and multiple 360$^{\circ}$ objects from observations of scenes at different timestamps. Recent works have used neural radiance fields to model 3D scenes and improved the quality of novel view synthesis, while few studies have focused on modeling the invisible or occluded parts of the training images. These under-reconstruction parts constrain both scene editing and rendering view selection, thereby limiting their utility for synthetic data generation for downstream tasks. Our basic idea is that, by observing the same set of objects in various arrangement, so that parts that are invisible in one scene may become visible in others. By fusing the visible parts from each scene, occlusion-free rendering of both background and foreground objects can be achieved.

We decompose the multi-scene fusion task into two main components: (1) objects/background segmentation and alignment, where we leverage point cloud-based methods tailored to our novel problem formulation; (2) radiance fields fusion, where we introduce \textit{visibility field} to quantify the visible information of radiance fields, and propose \textit{visibility-aware rendering} for the fusion of series of scenes, ultimately obtaining clean background and 360$^{\circ}$ object rendering. Comprehensive experiments were conducted on synthetic and real datasets, and the results demonstrate the effectiveness of our method.

\end{abstract}

\section{Introduction}
\label{sec:intro}

With the rapid development of deep learning technologies, learning-based robotics has become increasingly prevalent. For home robots, understanding indoor environments and everyday objects is crucial for their functioning. To achieve this, a substantial amount of high-quality annotated data (\eg, images/videos, object/scene labels, and object/camera poses) for daily scenarios is crucial. Given the labor-intensive nature of manual collection and annotation of such data, researchers have explored the use of 3D models capable of generating infinite annotated data via rendering for robot learning~\cite{dwibedi2017cut, tobin2017domain}. However, the manual creation of 3D assets demands considerable time and labor, making it impractical for diverse home environments and rapidly evolving everyday items. Hence, the efficient and low-cost acquisition of extensive 3D data for daily scenes has become a vital research topic.

Recent advancements in 3D reconstruction algorithms, represented by Neural Radiance Fields (NeRF)~\cite{Mildenhall2020NeRFRS}, enables the reconstruction of high-quality 3D scenes from captured videos. Nevertheless, directly applying these methods to robot learning tasks reveals certain shortcomings. The primary challenge with these methods is that, they mainly focus on reconstructing static scenes, whereas home robots encounter dynamic scenarios with objects placed in various poses. The abundance of diverse object configurations is crucial for downstream tasks in robotics, such as pose estimation and object detection. To generate a large amount of synthetic data with different object configurations through reconstructed scene, it becomes necessary to edit the scene content. However, even with the application of recent scene editing techniques~\cite{xu2022deforming, Yuan2022NeRFEditingGE}, rearranging objects on the reconstructed model introduces new challenges: parts not visible during capture (\eg, due to object occlusion) are under-reconstruction, and these regions may be exposed after scene editing, resulting in rendering artifacts that affect the quality of synthetic data (\eg, Fig.~\ref{fig:basic_idea}~(c)). 
This issue is not significant for items that inherently do not need to be moved (\eg, large furniture such as refrigerators) or for areas rarely visible in daily use (\eg, the bottom of a cup). However, for items that frequently change in location and orientation, such as toys or water bottles, this problem becomes more pronounced.

To address these challenges, this paper proposes a method for NeRF-based 3D reconstruction that recovers a clean background and multiple 360$^{\circ}$ foreground objects from series of captured videos at ones. The overview of our method is shown in Fig.~\ref{fig:teaser}.
Our approach goes beyond real-to-sim by reconstructing real-world scenes into simulated models. More importantly, due to the reconstructed models possessing complete clean background and unoccluded 360$^{\circ}$ object features, our method can generate rendering datasets with countless variations in object placements in the virtual world, thereby supporting a wide range of robot learning tasks in the real world.

We leverage NeRF to reconstruct series of scenes captured at different moments, and segment and match backgrounds and each foreground object in each scene. We employ a novel radiance field fusion algorithm to fuse each instance, yielding a clean background and multiple 360$^{\circ}$ foreground objects.
Specifically, given the volumetric nature of the radiance field, we propose \textit{visibility field}, a volumetric representation for quantifying the visibility in scenes. With the proposed visibility field, we compare the visibility of the corresponding part across series of scenes and fuse the parts with higher visibility to achieve clean background and multiple 360$^{\circ}$ objects rendering. We dub our proposed idea of visible part fusion as \textit{ViFu}. The basic idea of ViFu is shown in Fig.~\ref{fig:basic_idea}.
Furthermore, we leverage the series-of-scene setting and propose a method for segmenting objects and backgrounds by exploiting the differences in object placement across each scene. Our segmentation approach is based on the geometric differences w.r.t. clean backgrounds obtained via fusion, which is computationally efficient and simple, and does not require any pre-trained 3D segmentation model.

To verify the effectiveness of ViFu, we created several sets of synthetic scenes containing various objects. We observe that ViFu automatically and accurately segments the background and each object, and achieves pleasing recovery of clean backgrounds and free-view rendering of multiple 360$^{\circ}$ foreground objects. We also captured videos to create a set of real-world datasets, and the experimental results show that the proposed method also gives promising results for real-world scenes.

In summary, our main contributions are listed as follows:
\begin{itemize}
    \item We studied the under-reconstruction invisible parts of NeRF and introduced the setting of complementing the invisible parts by fusing series-of-scene information.
    \item We introduce \textit{visibility field}, a volumetric representation to quantify the visibility of scenes, and propose novel \textit{visibility-aware rendering}, which leverages the visibility field to achieve visible parts fusion of series of scenes.
    \item We created synthetic and real datasets to validate our idea, and the experimental results show the effectiveness of the proposed method.
\end{itemize}

\section{Related work}
\label{sec:related}

\paragraph{3D reconstruction}
3D reconstruction has been a crucial task in computer vision due to its widespread applications. In the early stages, classical photogrammetry formed the basis for 3D reconstruction, relying on image correspondences and geometric constraints to predict 3D structures from 2D images. While these methods are well-established, their performance is limited when dealing with texture-less or reflective surfaces, constraining their accuracy. Structure-from-Motion (SfM)~\cite{schoenberger2016sfm} reconstructs sparse point clouds from multi-view images, making it suitable for large-scale scenes. However, its sparse nature leads to the loss of geometric details. Multi-view stereo (MVS)~\cite{schoenberger2016mvs} refines SfM results by predicting dense point clouds, enhancing accuracy to some extent but still facing challenges related to detail preservation. Subsequently, the development of differentiable rendering allows the use of rendering errors to supervise the optimization of 3D models, facilitating the reconstruction of 3D scenes from 2D images. This idea has been applied to classical 3D representations, such as meshes~\cite{kato2018renderer,liu2019softras}, point clouds~\cite{insafutdinov2018unsupervised}, voxel grids~\cite{liu2020neural}, \etc. Meshes or point clouds, due to their explicit 3D representations, are easily utilized in downstream tasks. However, they still face challenges in reconstructing fine details when supervised solely using pixel loss.

Recently, implicit representations has received significant attention due to its detailed representation of the geometry and appearance of the scene~\cite{sitzmann2019srns, yariv2020multiview, Mildenhall2020NeRFRS}. The most representative work is neural radiance field (NeRF)~\cite{Mildenhall2020NeRFRS}, which uses neural networks to model the scene as a continuous mapping from position and view direction to radiance color and volume density, enabling geometric and appearance reconstruction and photorealistic novel view rendering. Several follow-up works have been proposed to improve the foundation of NeRF, enabling fast optimization~\cite{Yu2021PlenoxelsRF, Mller2022InstantNG}, appearance decoupling~\cite{Verbin2021RefNeRFSV}, dynamic scene modeling~\cite{Pumarola2020DNeRFNR, Park2020NerfiesDN}, and more. Nevertheless, these methods have limitations as they model the scene as a whole and do not allow for segmentation or editing of specific parts of the scene.

\paragraph{Manipulating NeRF for novel scene creation}
Utilizing the reconstructed NeRF for generating a diverse synthetic dataset is a natural idea. To achieve this, the manipulability of NeRF becomes crucial. This involves manipulations such as object/background segmentation, object movement/re-posing, object deletion/duplication, completion of under-reconstruction parts, etc. 
To achieve these objectives, compositional scene modeling methods~\cite{zhang2020nerf++, guo2020object, niemeyer2021giraffe, zhang2021editable, wu2022object} regard the entire scene as a mixture of background and foreground objects, facilitating object-level scene understanding; some methods encode semantic information into scenes, enabling feature-based object query or segmentation~\cite{Zhi:etal:ICCV2021, wang2022clip}. Another direction explores object-level manipulations on scene content, enabling editing to object appearance~\cite{Liu2021EditingCR, bao2023sine} or geometry~\cite{xu2022deforming, Yuan2022NeRFEditingGE}. These advancements have made notable progress in manipulating NeRF-based representations, however, our primary concern is that manipulating the original scenes (\eg, object movement) can inadvertently expose unseen parts and thus lead to artifacts.

\section{Method}
\begin{figure*}
\centering
\includegraphics[width=\textwidth]{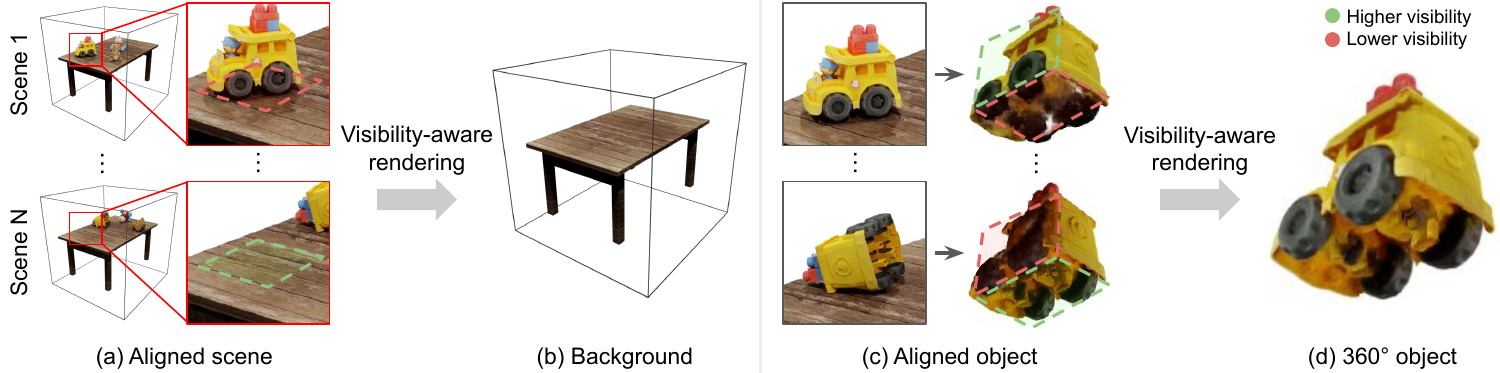}
\caption{\textbf{The basic idea of ViFu}. With pre-computed scene/objects alignment, we compare the visibility of the corresponding parts using the proposed visibility field, and fuse the higher visibility parts of each scene to form the clean background and multiple 360$^{\circ}$ objects. The details of visibility-aware rendering are shown in Fig.~\ref{fig:visibility_aware_rendering}.}
\label{fig:basic_idea}
\end{figure*}

Consider a static background and $M\geq1$ foreground objects that are placed in different locations and orientations resulting in $N\geq2$ different scenes (\eg, Fig.~\ref{fig:teaser}~(a)). For each scene, we capture $L_i$ multi-view images $\{ \mathcal{I}_l \}$ and run the structure-from-motion method independently for each scene to obtain the camera parameters (intrinsics and extrinsics) $\{\mathcal{C}_l\}$, where $i\in\{1,...,N\}$ denotes scene index and $l\in\{1,...,L_i\}$ denotes camera index of scene $i$.
From the calibrated multi-view images, we optimize neural radiance fields (NeRF) $\{\mathcal{S}_i\}$ for each scene. Radiance field is an implicit scene representation that maps spatial position $\mathbf{x} \in \mathbb{R}^3$ and view direction $\mathbf{d} \in \mathbb{S}^2$ to radiance color $\mathbf{c}=(r,g,b)$ and volume density $\sigma$ as $\mathcal{S}: (\mathbf{x}, \mathbf{d}) \mapsto (\mathbf{c}, \sigma)$.

Our method takes $N$ optimized radiance fields $\{\mathcal{S}_i\}$ as input, automatically splits the scenes into a static background and $M$ foreground objects, and recovers a non-occluded background and multiple 360$^{\circ}$ objects that can be seen from arbitrary view point.

\paragraph{Terminology} We denote the parts that exhibit minimal variation in locations/orientations over time as \textit{background} (\eg, table or floor), while those that undergo temporal changes are termed \textit{foreground objects} (\eg, items on the desk). Additionally, the term \textit{scene} refers to the union of background and foreground(s), with \textit{series of scenes} representing the collection of scenes at multiple timestamps.

\paragraph{Assumption} \label{par:assump} For our problem formulation, we make the following two ideal assumptions: (1) diverse object \textit{locations}: this ensures visibility of the background, implying that every part of the background is observable in at least one scene (\ie, no permanently occluded regions); this also facilitates the segmentation of foreground objects, as will be introduced in Sec.~\ref{sec:segmentation-and-alignment}. (2) diverse object \textit{orientations}: this guarantees that every part of the object's surface is observable in at least one scene (\eg, no permanently facing-down surfaces).

The assumptions are natural for household objects in everyday scenes: static objects that remain unchanged, such as refrigerators or tables, are considered part of the background; while objects that are frequently moved, such as the toys in Fig.~\ref{fig:teaser}, are treated as foreground objects.

\subsection{Method overview}
Our objective is to perform background/foreground segmentation from series of scenes and obtain a clean background and multiple 360$^{\circ}$ objects via fusion. In the general context of 3D modeling, this process can be divided into two main steps: the first involves \textit{internal scene} reasoning, specifically the segmentation of background/foreground within each scene; the second entails \textit{inter-scene} reasoning, which involves matching the segmented background and individual objects among different scenes (\ie, pose alignment for background and foreground objects), and subsequently accomplishing the final fusion.

In the following section, we introduce our solutions, specifically tailored for the recent 3D representation of the radiance field. To be more precise, we leverage a point cloud-based approach to perform scene segmentation and alignment (Sec.~\ref{sec:segmentation-and-alignment}), and introduce a novel measure for quantifying the visibility of the radiance fields (Sec.~\ref{sec:visibility_field}), which is used in the proposed scene fusion method (Sec.~\ref{sec:visibility_aware_rendering}).

\subsection{Object segmentation and alignment}
\label{sec:segmentation-and-alignment}
The first step involves background/foreground segmentation and obtaining the relative poses of foreground objects and background within each scene. This allows us to align them to their respective common coordinate systems, which are utilized for subsequent fusion purposes (See Fig.~\ref{fig:basic_idea}). For the segmentation and alignment of the radiance fields, we found that existing point cloud-based methods already yield satisfactory results. For simplicity, we introduce here the minimal segmentation and alignment techniques below; however, other more advanced alternatives can also be employed.

We provide a high-level overview of the entire process.
First, we employ Marching Cubes~\cite{lorensen1987marching} to convert the radiance field of each scene into a mesh, from which we extract point clouds by surface sampling. 
While the placement of individual foreground objects may vary, a substantial overlap of point clouds belonging to the static background is sufficient for achieving inter-scene pose alignment through point cloud registration algorithms. 
Based on the derived relative poses, we utilize the method outlined in Sec.~\ref{sec:visibility_aware_rendering} to obtain the fused clean background, and from which we similarly extract the point cloud corresponding to the background.
By comparing the differences between the point clouds of each scene and the clean background, we can obtain all the point clouds that belong to the foreground objects. Subsequently, a point cloud clustering algorithm (\eg, \cite{ester1996density}) allows us to obtain point clouds that belong to each individual foreground object separately.
Finally, for each foreground object across scenes, the Hungarian matching algorithm~\cite{kuhn1955hungarian} and point cloud registration techniques are used to determine their correspondences and relative poses $\{T_{i,j}\}$. Here $j\in\{1,...,M\}$ denotes the object index.

\subsection{Visibility field: quantifying visibility in radiance field}
\label{sec:visibility_field}
Visibility is an important measure to utilize the \textit{visible part} information across scenes. 
To quantify the visibility information in the radiance field, we propose \textit{visibility field}, a volumetric representation that maps a 3D position to a scalar-valued visibility:
\begin{equation}
    \label{eq:visibility_field_definition}
    v = v(\mathbf{x}): \mathbb{R}^3 \rightarrow [0, 1] .
\end{equation}
The proposed visibility $v(\mathbf{x}) \in [0, 1]$ is defined as the proportion of cameras that can observe point $\mathbf{x}$ among all training cameras. 
Formally, we say that $\mathbf{x}$ can be observed by the camera $\mathcal{C}_l$ means that (1) the projection of $\mathbf{x}$ falls within the interior of the image plane and (2) there is no occlusion between $\mathbf{x}$ and the camera position $\mathbf{o}_l \in \mathbb{R}^3$.
For the condition (2), we use the pseudo-depth of the radiance field to determine whether there is occlusion. Specifically, we cast a ray from the camera position $\mathbf{o}_l$ to $\mathbf{x}$ and compute the pseudo-depth $\hat{d}_l$ by volume rendering, and then compare it with the distance from the camera position to the point $d_l = \|\mathbf{x} - \mathbf{o}_l\|$. 
For camera $\mathcal{C}_l$, we use a binary-valued function $V_l(\mathbf{x}) \in \{0, 1\}$ to denote whether $\mathbf{x}$ can be observed by that camera. If $d_l < \hat{d}_l$, this means that $\mathbf{x}$ is between the object surface and the camera position, thus there is no occlusion, \ie, $V_l(\mathbf{x}) = 1$, otherwise $V_l(\mathbf{x}) = 0$.
Considering all training cameras, the visibility of position $\mathbf{x}$ can be computed as:
\begin{equation}
    \label{eq:visibility_field_sum}
    v(\mathbf{x}) = \frac{1}{L}\sum_{l=1}^{L} V_l(\mathbf{x}) .
\end{equation}
Note that the visibility field is independent for each scene and we compute it for all scenes.

\subsection{Visibility-aware rendering}
\label{sec:visibility_aware_rendering}

 \begin{figure}[]{}
    \includegraphics[width=\linewidth]{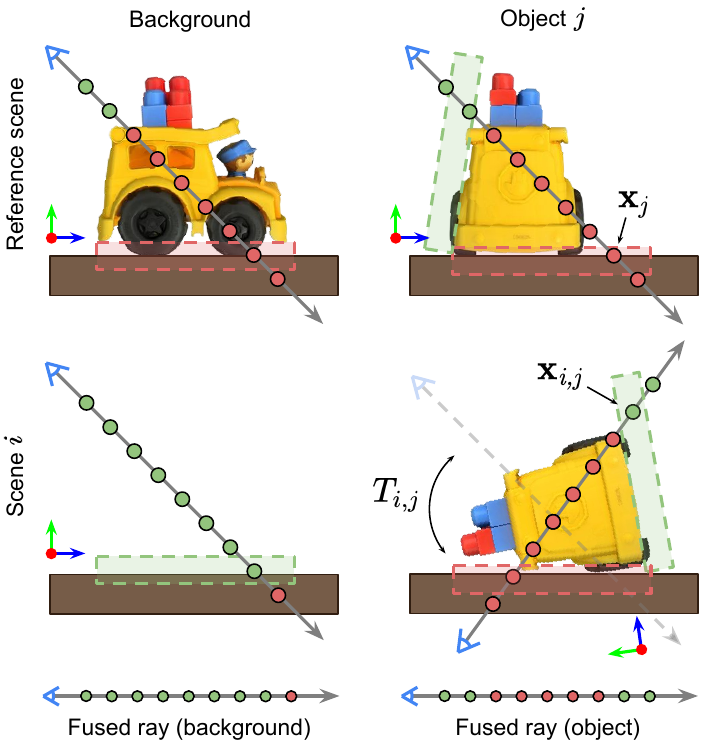}
 \caption{Illustration of visibility-aware rendering in 2D. The colors correspond to higher/lower visibility as shown in Fig.~\ref{fig:basic_idea}.
}
\label{fig:visibility_aware_rendering}
\vspace{-1.2em}
\end{figure}

\label{sec:visibility-aware-rendering-bg}
We propose \textit{visibility-aware rendering}, a method that obtains occlusion-free rendering by comparing the visibility across series of scenes.
We take the rendering of the clean background to explain its basic idea (Fig.~\ref{fig:visibility_aware_rendering}~(Left)).

\paragraph{Background}
The first step in comparing scenes is to set them under the same coordinates.
Recall that we have obtained the relative pose between series of scenes through point cloud registration in Sec.~\ref{sec:segmentation-and-alignment}. Without loss of generality, we take the first scene $(i=1)$ as a reference and align the scenes $i=2,...,N$ to the coordinate system of the first scene.
Given that all scenes are aligned to the reference scene, we introduce visibility-aware rendering for the background. An illustration is shown in Fig.~\ref{fig:visibility_aware_rendering} (Left). For the sample point $\mathbf{x}$ in volume rendering, the proposed visibility-aware rendering, in addition to color and density, also computes the visibility of the sample point in each scene, \ie, $\{\mathbf{c}_i(\mathbf{x})\}$, $\{\sigma_i(\mathbf{x})\}$ and $\{v_i(\mathbf{x})\}$.
The idea of visibility-aware rendering is simple: blend the color and density in each scene according to visibility. Using $w_i$ which satisfy $\sum_i w_i = 1$ to denote the weight of each scene, blended radiance color and volume density can be written as:
\begin{equation}
\label{eq:blended-color-density}
    \hat{\mathbf{c}}(\mathbf{x}) = \sum_{i=1}^{N} w_i(\mathbf{x}) \mathbf{c}_i(\mathbf{x}), \
    \hat{\sigma}(\mathbf{x}) = \sum_{i=1}^{N} w_i(\mathbf{x}) \sigma_i(\mathbf{x}), 
\end{equation}
where $w_i$ is a weight function calculated from visibility that satisfies $\sum_i w_i = 1$:
\begin{equation}
\label{eq:weights-from-visibility}
w_i(\mathbf{x}) = \frac{v_i^p(\mathbf{x})}{\sum_{i=1}^N v_i^p(\mathbf{x})}.
\end{equation}
Here $p$ is a hyper-parameter that controls the weights, the larger $p$ is, the greater the contribution of the scene with the highest visibility; and when $p \rightarrow \infty$, the above is equivalent to the max-selection function. For simplicity, Fig.~\ref{fig:visibility_aware_rendering} shows the case based on max-selection.

The motivation behind the above calculation is to select the parts with less occlusion (\ie, higher visibility) in each scene, and fuse them into the final scene. As a result, volume rendering of the blended radiance color and volume density obtained from Eq.~(\ref{eq:blended-color-density}) yields a clean background, as shown in Fig.~\ref{fig:basic_idea}~(b).

\paragraph{Foreground objects}
The core idea of visibility-aware rendering for 360$^{\circ}$ objects is basically the same as that for background.
Similarly, we take the coordinate systems of the foreground objects in scene $i=1$ as a reference. For foreground object $j$, we denote the position and view direction of the sampled point under the reference coordinate system as $\mathbf{x}_j$, $\mathbf{d}_j$, respectively. 
For scenes of $i\geq2$, we use the computed object poses to calculate the corresponding positions $\mathbf{x}_{i,j}$ and view directions $\mathbf{d}_{i,j}$ in each scene as:
\begin{equation}
\mathbf{x}_{i,j} = R_{i,j}\mathbf{x}_j + t_{i,j}, \ \mathbf{d}_{i,j} = R_{i,j}\mathbf{d}_j, 
\end{equation}
where $R_{i,j}$ and $t_{i,j}$ are rotation and translation terms of object poses $T_{i,j}\in \mathrm{SE(3)} $ obtained from Sec.~\ref{sec:segmentation-and-alignment}.
Here, $\mathbf{x}_{i,j}$ in fact represents the corresponding point of $\mathbf{x}_j$ in the coordinate system of scene $i$, as shown in Fig.~\ref{fig:visibility_aware_rendering}~(Left).
Then, the blended radiance color and volume density of Eq.~(\ref{eq:blended-color-density}) for foreground object rendering can be rewritten as:
\begin{equation}
\label{eq:blended-color-density-object}
    \tilde{\mathbf{c}}(\mathbf{x}_{j}) = \sum_{i=1}^{N} w_i(\mathbf{x}_{i,j}) \mathbf{c}_i(\mathbf{x}_{i,j}), \
    \tilde{\sigma}(\mathbf{x}_{j}) = \sum_{i=1}^{N} w_i(\mathbf{x}_{i,j}) \sigma_i(\mathbf{x}_{i,j}).
\end{equation}
Volume rendering the fusion results obtained from Eq.~(\ref{eq:blended-color-density-object}) yields occlusion-free 360$^\circ$ foreground objects, as shown in Fig.~\ref{fig:basic_idea}~(d).

Our proposed visibility-aware rendering, despite its simplicity, reasonably achieves the visible part fusion of radiance fields. It's noteworthy that our method share the same paradigm for both background/foreground parts, accomplishing the reconstruction of a clean background and 360$^\circ$ foreground objects.

\section{Experiments}

\subsection{Datasets}

\begin{figure*}
    \centering
    \includegraphics[width=\textwidth]{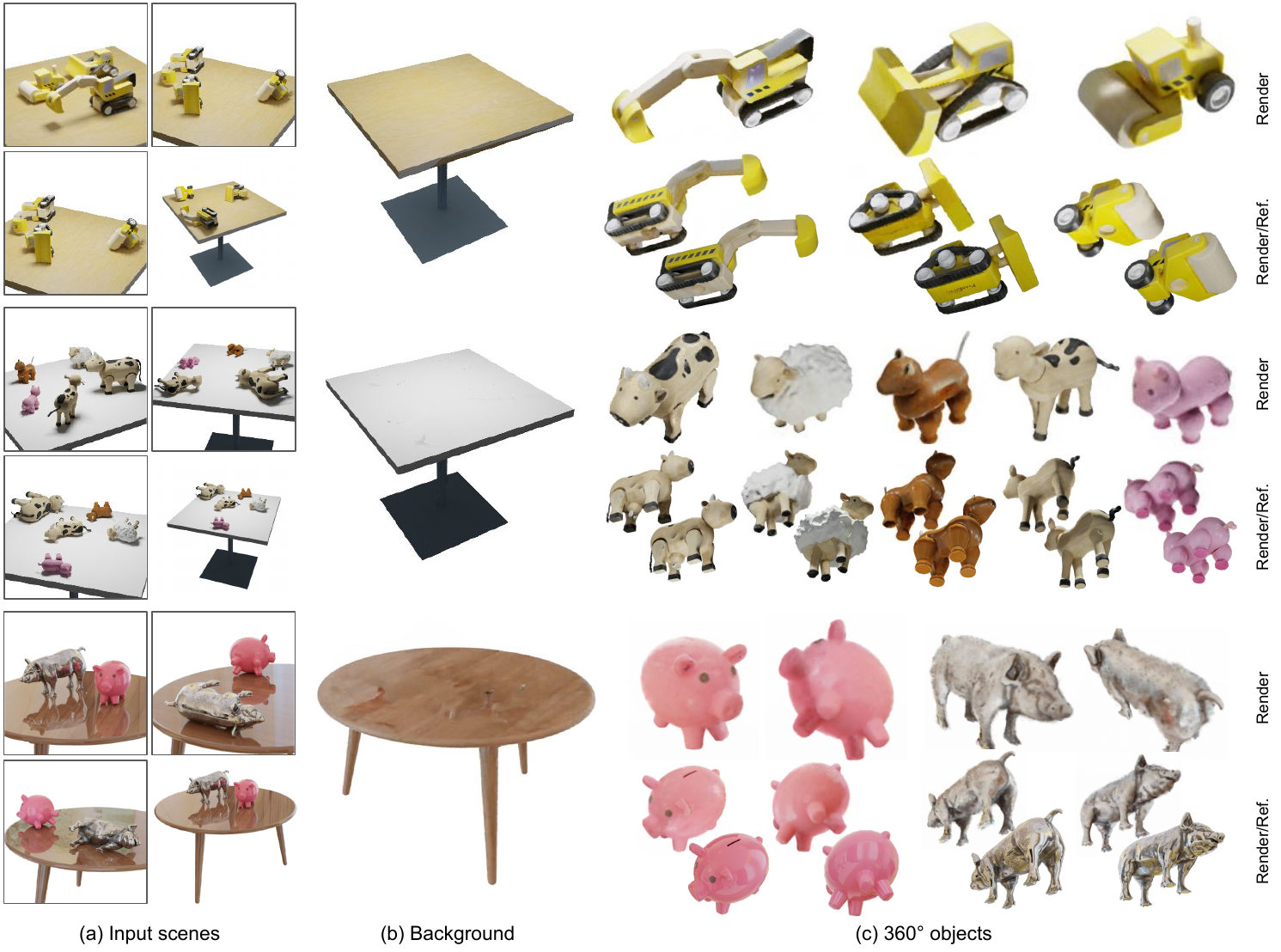}
    \caption{\textbf{Results on Blender synthetic datasets}. For pairwise comparisons of foreground objects, the top-left image shows the rendering result of the proposed method, while the bottom-right image shows the reference image (ground truth).}
    \label{fig:results_synthetic}
\end{figure*}

\paragraph{Blender synthetic datasets}
We created synthetic datasets using Blender~\cite{blender}. The tables used as background are taken from free 3D models available online. For the foreground objects, we use 3D models from Google Scanned Objects dataset~\cite{Downs2022GoogleSO}, which contains 360$^{\circ}$ scans of common household objects. 
We created $N=3$ sets of scenes, in which foreground objects are under difference placement to ensure that every part of the table and object surfaces is visible in at least one scene.
We applied different lighting conditions (uniform light, spotlight, \etc.) to test the effectiveness of our method in different environments.
We randomly sample camera positions on the hemisphere and render $L=100$ images for the radiance field optimization.
Examples of the synthetic scenes are shown in Fig.~\ref{fig:results_synthetic}~(a).

\begin{figure*}
    \centering
    \includegraphics[width=\textwidth]{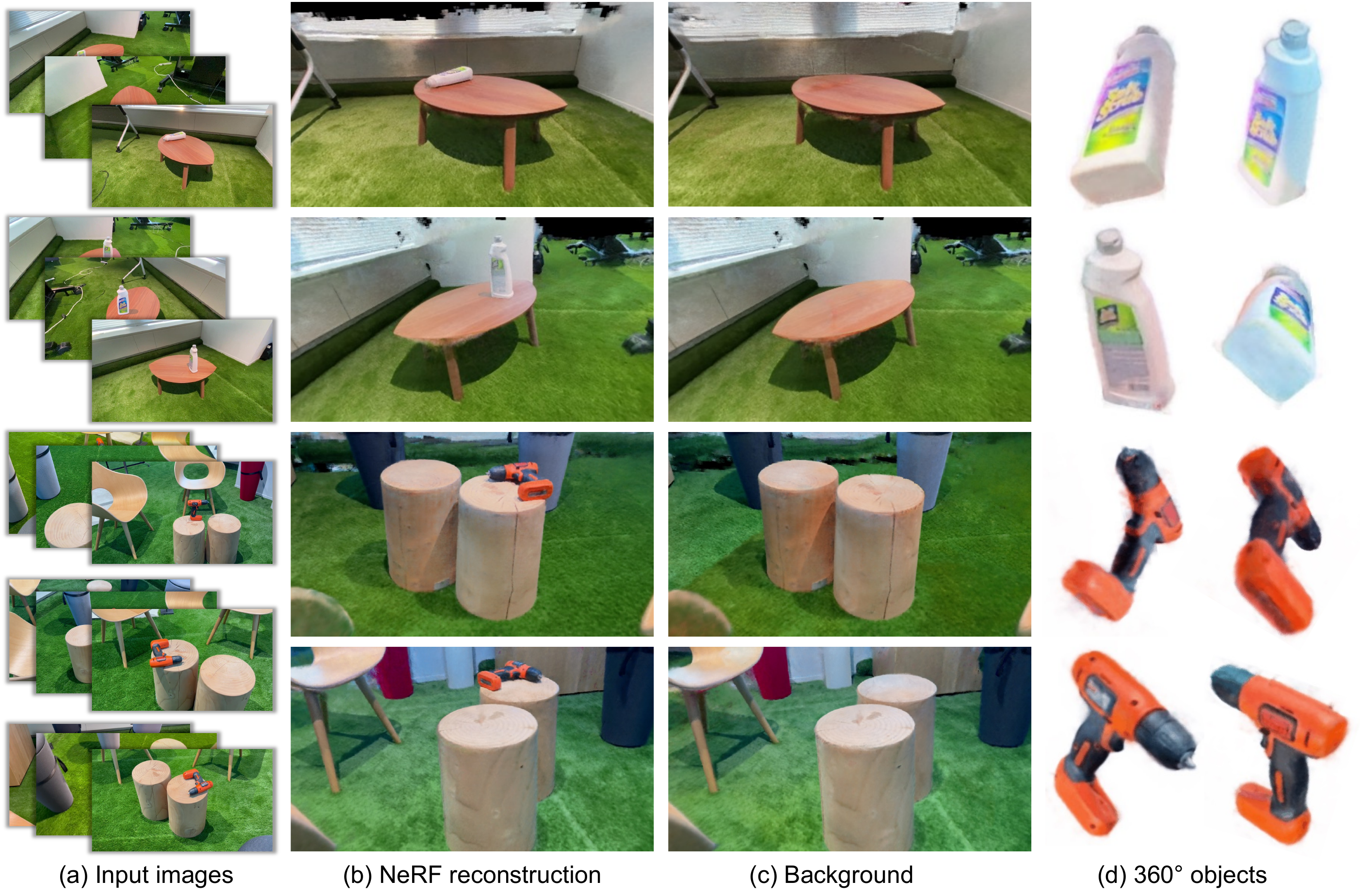}
    \caption{\textbf{Results on real capture datasets}. (c) and (d) are obtained using the proposed method.}
    \label{fig:results_real}
    \vspace{-1.2em}
\end{figure*}

\paragraph{Real capture datasets}
We created real-world capture datasets to demonstrate the effectiveness of our approach on real datasets. 
We utilized YCB objects~\cite{calli2015ycb} and created $N=2$ (for bleach cleanser) or $N=3$ (for power drill) scenes by placing objects in different configurations. For each scene, we captured a video around it and extracted 60-80 frames, then applied COLMAP~\cite{schoenberger2016sfm, schoenberger2016mvs} to obtain the corresponding camera parameters registration.
Examples of the real capture scenes are shown in Fig.~\ref{fig:results_real}~(a).

\subsection{Results}

We show the qualitative results of Blender synthetic datasets and real capture datasets on Fig.~\ref{fig:results_synthetic} and Fig.~\ref{fig:results_real}, respectively. With multiple input scenes, our method can automatically recover a clean background and multiple 360$^{\circ}$ foreground objects. Noticeably, to the best of our knowledge, there are currently no previous works using the problem setting of series of scenes, nor addressing the issue of under-reconstruction in NeRF literature. Therefore, direct comparison between our method and existing methods becomes challenging. In turn, we conduct an in-depth ablation study on the proposed method in the subsequent section.

\subsection{Ablation studies}
\label{sec:ablation}

\paragraph{Impact of light conditions}
\begin{figure}[h]
\centering
\includegraphics[width=\linewidth]{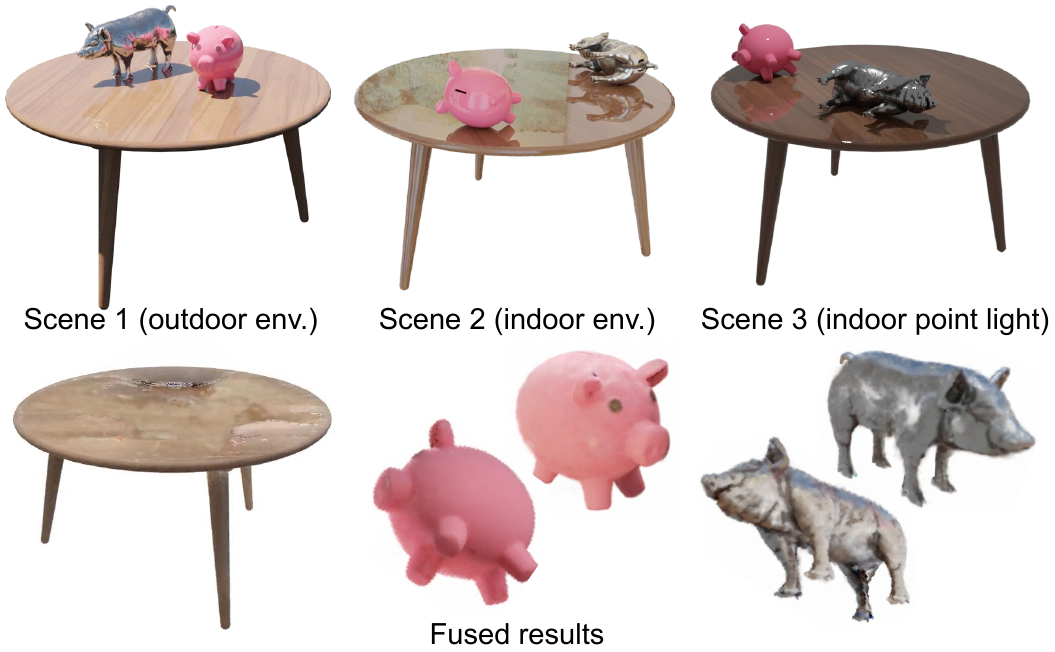}
\caption{Ablation on different light conditions.}
\label{fig:ablation_light}
\vspace{-1.5em}
\end{figure}
We created scenes under three distinct lighting conditions: outdoor environment mapping, indoor environment mapping, and a single point light source. (See Fig.~\ref{fig:ablation_light}) For the background, despite obtaining an acceptable clean background, there exists a certain degree of artifacts due to the presence of different shadows or reflections. For objects, certain discontinuities arise due to abrupt changes in lighting conditions or the inherent glossiness of the objects (\eg, pink pig). Additionally, the fused results show a lack of glossiness, suggesting that even for significantly different lighting conditions or glossy objects, our fusion method can neutralize the view-dependent term, yielding appearances close to diffuse colors, which is typically desirable in the context of 360$^\circ$ object reconstruction.
\paragraph{Impact of weight function}
\begin{figure}[h]
\centering
\includegraphics[width=\linewidth]{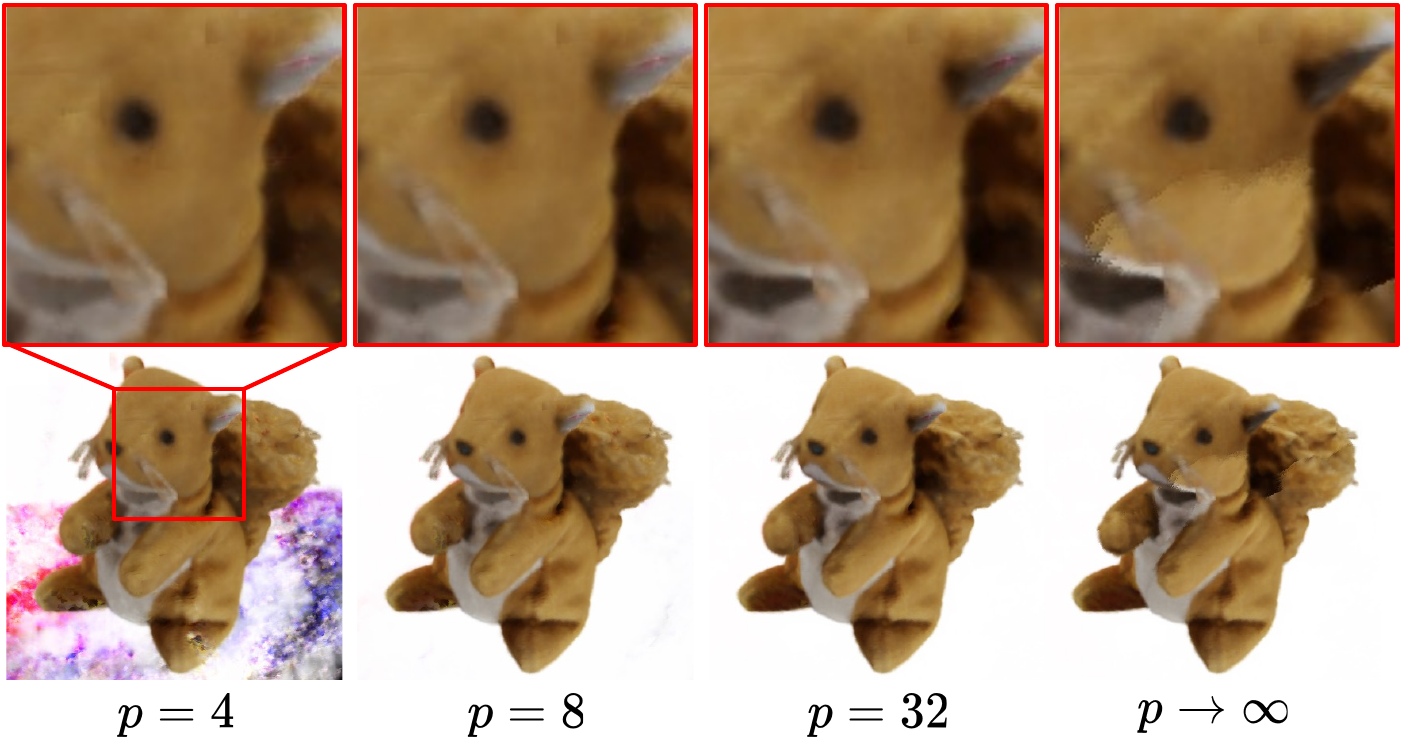}
\caption{Ablation on weight function. $p$ is the parameter controls the weights in Eq.~(\ref{eq:weights-from-visibility}).}
\label{fig:ablation_weight}
\vspace{-1.5em}
\end{figure}
We study the impact of the hyper-parameter $p$ (exponent of visibility in weight function Eq.~(\ref{eq:weights-from-visibility})) (See Fig.~\ref{fig:ablation_weight}). We observe that when $p$ is relatively small (\ie, $p=4$), the results tend to blend color and density more smoothly for each scene. The appearance changes smoothly for foreground objects, however, it also blends the background and non-background (\ie, empty space) parts around them, resulting in a cloud-like artifact. When $p \rightarrow \infty$, visibility aware rendering selects the color and density of the scene with the highest visibility as the result of the fusion, and such a max-selection brings discontinuous changes, resulting in sharp changes in the appearance. We observe that $p=8\sim32$ is the appropriate value to obtain continuous appearance interpolation without cloud-like artifacts.

\paragraph{Impact of the number of scenes}
\begin{figure}[h]
\centering
\includegraphics[width=\linewidth]{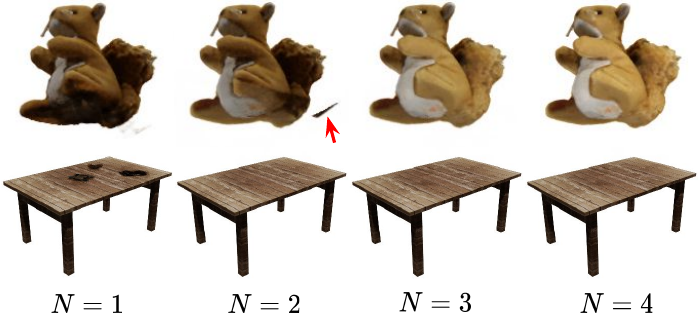}
\caption{Ablation on the number of scenes used for fusion.}
\label{fig:ablation_num}
\vspace{-1.5em}
\end{figure}
Right image shows the impact on the results for different numbers of scenes $N$ (See Fig~\ref{fig:ablation_num}). 
For $N=1$, we manually compute the bounding boxes for the background and foreground objects from the point cloud and rendered only the original scene within them. In this case, the invisible parts are not optimized, leading to artifacts in the rendering results. For $N>2$, we can observe that the proposed ViFu can recover a clean background and multiple 360$^{\circ}$ objects from series-of-scene observations. 

It is noteworthy that, as the number of scenes $N$ increases, the rendered results appear to become brighter. We speculate that this is due to sufficient lighting generally implying less occlusion, which means higher visibility and thus the corresponding parts are fused into the final output with higher weight. Based on this observation, we assume that as the number of scenes and the variety of object poses increase, the rendering results of objects will be close to those rendered in a 360$^{\circ}$ spherical lighting environment.

Empirically, we observe that for some objects, artifacts appear when $N=2$ (red arrow in the figure). 
We attribute this to the difficulty in accurately segmenting the foreground object if a certain part is in contact with the background part in both two scenes, making it hard to determine whether it belongs to the foreground object or background.
A simple solution is to expose the part of the common contact when placing objects in the third scene. Although an adhoc placement may achieve the plausible rendering at $N=2$ (as the cleanser scenes in Fig.~\ref{fig:results_real}), 
we observe that $N=3$ scenes can achieve reasonable segmentation in most cases and is therefore a recommended choice.

\paragraph{Impact of variations in object placement}
To validate the robustness of our approach to variations in object placement, we extended our evaluation beyond the 3 original scenes presented in Fig.~\ref{fig:teaser}. We created an additional 5 scenes, where objects were randomly placed. We randomly selected 3 from 8 scenes for each fusion experiment. We have the following observations: under the assumptions~\ref{par:assump}, our method consistently produced satisfactory results. However, 
some difficulties arise:
(1) when the orientations of foreground objects in the selected scenes are highly repetitive (\eg, bottoms consistently facing downward and thus not observable), artifacts are still present in the rendered regions that lacked sufficient observation. This issue arises because our method relies on fusing information from the available scenes and thus cannot predict unseen part.
(2) when foreground objects are placed very closely within a scene, our use of a naive point cloud segmentation approach may potentially fail, leading to misalignment and bad fusion results. Effective segmentation of closely spaced objects typically requires prior knowledge of the objects. Incorporating pre-trained point cloud segmentation models or segmentation masks as additional information can assist in segmenting challenging objects, thereby facilitating successful scene fusion.

\section{Limitations and Future work}
There are a few limitations that need to be addressed in future work:
(1) First, our method does not explicitly consider the lighting condition. For static background, as the lighting conditions are basically the same, reasonable rendering results can be obtained. 
The above ablation study for light conditions demonstrates that our proposed weighted fusion method can mitigate the impact of certain lighting variations to some extent. However, the rendering results of objects under extreme lighting conditions may still be unsatisfactory (\eg, the fusion result of ``construction vehicles'' at the top of Fig.~\ref{fig:results_synthetic} shows an abrupt change in appearance, where spot light illumination is used). Incorporating some of the current approaches for disentangling light conditions might be a promising direction for future work.
(2) Second, our fusion method assumes that we can obtain accurate scene segmentation and pose alignment. In most cases, the aforementioned point cloud-based approach can achieve sufficiently accurate segmentation and alignment. However, some challenging scenarios may arise, such as failures in segmentation due to close object placement (as mentioned in Sec.~\ref{sec:ablation}), or failures in pose alignment due to oversimple object shapes. 
However, the essence of these problems can all be viewed as fundamentally challenging issues in point cloud segmentation or registration, which has been a longstanding challenging problem in the field of computer vision.
For these special cases, using additional masks or richer point cloud features (\eg, color information) might help mitigate the aforementioned challenges.

\section{Conclusion}
We have presented ViFu, a method for recovering clean background and multiple 360$^{\circ}$ foreground objects from observations of scenes at different timestamps. We leverage point cloud-based approaches to achieve background and foreground alignment and use the difference between scenes to obtain a background/foreground segmentation. We propose visibility field, a volumetric representation to quantify the visibility of a scene, and introduce visibility-aware rendering to fuse the more visible parts of series of scenes. 
Our experiments on both synthetic and real datasets demonstrate the effectiveness of our approach. While our approach is the first to focus on radiance fields for scenes at different timestamps, there are some remaining issues, such as not considering lighting conditions, which we plan to address in future work.

\appendix
\subsection{Implementation details}

\paragraph{Radiance field representation}
We implement our method using Instant-NGP~\cite{Mller2022InstantNG}, a state-of-the-art radiance field representation with fast optimization, where each scene can be optimized within a few minutes in order of magnitude.

\paragraph{Visibility field representation}
Given the recent advances in using an explicit grid to represent neural field, we also use an explicit grid to model the proposed visibility field.
Specifically, for the target region we discretize the space into $64^3$ grid points, each of which holds the computed visibility of that point. Trilinear interpolation is used to compute the visibility of any point in the continuous 3D space.
We observe that, in the original formulation of visibility Eq.~(\ref{eq:visibility_field_sum}), the discontinuous nature of $V(x)$ (\ie, the visibility may abruptly change from visible to invisible near the surface of an object) may leads to discontinuous scene fusion and poor rendering results. Therefore, we apply smoothing to the computed visibility field. Please refer to the following Sec.~\ref{sec:visibility-smoothing} for more discussion.

\subsection{Scene poses registration}
\label{app:scene_pose_registration}
For $N$ scenes, we run COLMAP~\cite{schoenberger2016sfm, schoenberger2016mvs} independently for each scene. As a result, the camera poses and optimized radiance fields for each scene are expressed in different coordinate systems. To enable comparison of radiance fields across scenes, we introduce a method for aligning the scenes to a reference scene. Without loss of generality, we chose the first scene as the reference scene. Due to the lack of scale information in COLMAP, the alignment between scenes is a $\mathrm{Sim(3)}$ registration problem.

Given the scene point cloud $\{\mathcal{P}_i\}$ computed from each radiance fields, the scene alignment task can be formulated as finding an appropriate transformation $Q_i\in \mathrm{Sim(3)}$ for the point cloud $\mathcal{P}_i$ that minimizes the error $||\mathcal{P}_{\mathrm{ref}} - Q_i \mathcal{P}_i||$ where $\mathcal{P}_{\mathrm{ref}}$ is the point cloud of the reference scene. Note that although the foreground object placements varies across scenes, we assume that the static background provides sufficient information for obtaining reasonable scene registration results.

However, we have observed that relying solely on point cloud information can be challenging since the geometric information of the background may not be unique enough, leading to unstable registration results. Hence, we propose to leverage the available multi-view image information to facilitate the registration process.
Specifically, for scene $i$, we consider the multi-view images $\{\mathcal{I}_l\}_i$ and the 3D position $\{\mathbf{q}_l\}_i$ of its registered camera, where $\mathbf{q}_l \in \mathbb{R}^3$ is expressed in the coordinate system of scene $i$. Using the COLMAP registration of the reference scene, we register images $\{\mathcal{I}_l\}_i$ to the reference scene to obtain the camera position $\{\mathbf{q}_l\}'_i$ expressed in the reference coordinate system. Although the object placements in scene $i$ differ from those in the reference scene, the RGB information in the background provides sufficient information to complete the registration. 
Here, since $\{\mathbf{q}_l\}_i$ and $\{\mathbf{q}_l\}'_i$ represent the same camera position expressed in the coordinate systems of scene $i$ and the reference scene, respectively, given the relative pose between the scenes $Q_i$, it should satisfy that $\{\mathbf{q}_l\}'_i = Q_i \{\mathbf{q}_l\}_i$ for all $l$. Therefore, we solve for the transformation $Q_i \in \mathrm{Sim(3)}$ that minimizes the registration error between camera positions, given by
\begin{equation}
    \tilde{Q}_i = \argmin_{Q_i} \sum_{l} ||\{\mathbf{q}_l\}'_i - Q_i \{\mathbf{q}_l\}_i||.
\end{equation}
We use RANSAC~\cite{Fischler1981RandomSC} to solve for $\tilde{Q}_i$. After obtaining the transformation $\tilde{Q}_i$ we use it as the initial pose for scene point clouds registration. First, we align the scene point cloud with $\tilde{Q}_i$, and then refine the alignment between $\mathcal{P}_i$ and $\mathcal{P}_{\mathrm{ref}}$ using the Iterative Closest Point (ICP) algorithm~\cite{Rusinkiewicz2001EfficientVO}. This yields the final registered pose $Q_i$ for the scene.

\subsection{Object poses registration}
For the alignment of object point clouds, we first employ a global registration method to obtain an approximate alignment, followed by refinement using local alignment methods. Since the scale between different scenes has already been determined during scene registration as described in~\ref{app:scene_pose_registration}, object alignment here is an $\mathrm{SE(3)}$ registration problem. Specifically, as a global registration, we first extract FPFH features~\cite{Rusu2009FastPF} from the point cloud, and then use RANSAC~\cite{Fischler1981RandomSC} to obtain rough object poses estimation $T'_{i, j}$. Then, we use point-to-plane ICP~\cite{Arun1987LeastSquaresFO} to obtain the refined pose $T_{i, j}$. We denote the two-step procedure as a function \texttt{register}$(\cdot, \cdot)$, which takes two point clouds $\mathcal{P}_1$ and $\mathcal{P}_2$ as inputs and outputs their relative pose $Q \in \mathrm{SE(3)}$ and the registration fitness score $s\in [0, 1]$. The computation can be achieved using Open3D~\cite{Zhou2018Open3DAM} with just a few lines of code. 

As mentioned in the main paper, the correspondence between objects across different scenes is unknown. Therefore, we now describe how to simultaneously solve object matching and pose alignment problem. The pseudo-code is shown in Algorithm.~\ref{alg:matching}.
Considering the $j$ objects in the reference scene, we use $T_{i, j}$ to denote the relative pose of object $j$ in scene $i$ (\ie, the term we want to find). Note that we take the first scene as the reference scene, therefore $T_{1, j}=\mathbb{I}^{4\times 4}$ is the identity matrix.

We begin by aligning objects between the reference scene and scene $i$. For the $j$-th object in the reference scene, we pair it with $j'$-th object in scene $i$, align them, and calculate the aligned pose $Q$ with the corresponding fitness score $s$. This process is computed for all $j, j' \in \{1, \cdots, M\}$, resulting in a total of $M\times M$ relative poses and corresponding fitness values, which are then recorded in $\{\mathbf{Q}\}$ and $S$, respectively. Next, we use bipartite matching on the obtained cost matrix $S \in [0, 1]^{M \times M}$ to calculate the object matching that maximizes the overall fitness, and finally obtain the optimal correspondences and relative poses. 

By repeating the above process for all $i$, we can obtain the poses of objects in all scenes relative to the reference scene.

\begin{algorithm}
\caption{Object matching and alignment}\label{alg:matching}
\SetKwInOut{Input}{Input}\SetKwInOut{Output}{Output}
\Input{object point clouds $\{\tilde{\mathcal{P}}_{i,j}\}$}
\Output{object poses $\{T_{i, j}\}$}
$T_{1, j} \gets \mathbb{I}^{4\times4}$ \\
\For{$i = 2,\cdots,N$}{
    $ S \gets \mathbf{0}^{M\times M}$ \\
    $\{ \mathbf{Q} \} \gets \{\mathbb{I}^{4\times 4}\}_{j=1,\cdots,M;\ j'=1,\cdots,M}$ \\
    \For{$j = 1,\cdots,M$}{
        \For{$j' = 1,\cdots,M$}{ 
            $(Q, s) \gets $ \texttt{register}$(\tilde{\mathcal{P}}_{1,j}, \tilde{\mathcal{P}}_{i,j'})$ \\
            $S_{jj'} \gets s \in [0, 1]$ \\
            $\{\mathbf{Q}\}_{j, j'} \gets Q \in \mathrm{SE(3)}$ \\
        }    
    }
$index \gets$ \texttt{bipartite\_matching}$(S)$ \\
\For{$j = 1,\cdots,M$}{ 
    $T_{i, j} \gets \{\mathbf{Q}\}_{j, index[j]}$ \\
}
}
\Return{$\{T_{i, j}\}$}
\end{algorithm}

\subsection{Fused scene labels visualization}
\begin{figure*}
    \centering
    \includegraphics[width=\linewidth]{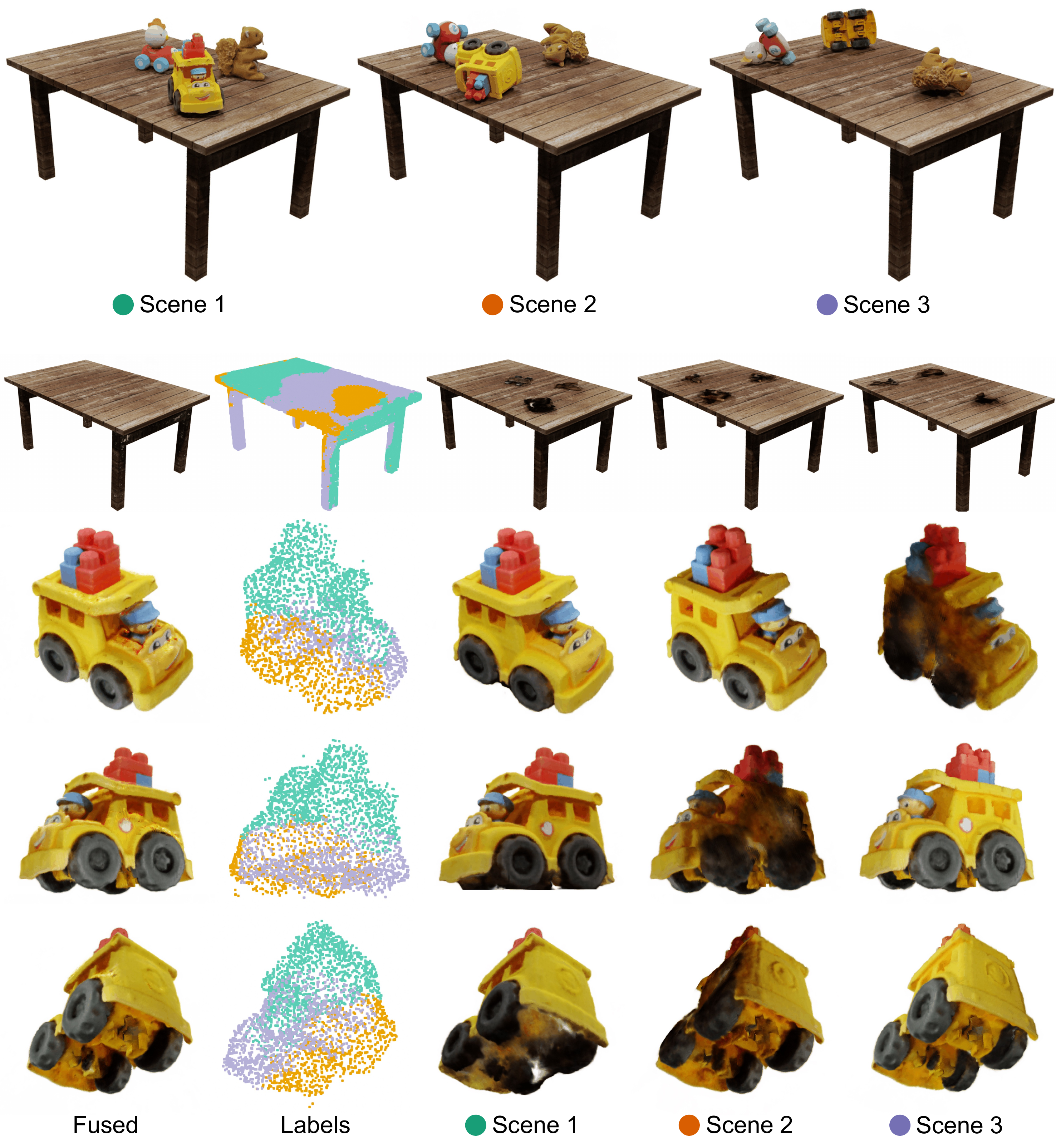}
    \caption{\textbf{Visualization of scene labels.} We show the correspondence between different parts of the fused scene and the original scenes. (Top) three original input scenes. (Bottom) background and foreground objects with fused scene, the point cloud labels, and the segmented original scenes. We can observe that, artifacts can appear when rendered from certain viewpoints due to occlusions in the input scenes, while our method accurately segments the parts of the space with higher visibility based on the proposed visibility field, thus synthesizing scenes without occlusion.}
    \label{fig:ablation_labels}
\end{figure*}
As our method fuses multiple scenes into one, we show in Fig.~\ref{fig:ablation_labels} which parts of the fused scene come from which original scene. 
Specifically, for the obtained point clouds of background and foreground objects, we compute the visibility of each point at their corresponding positions in each original scene, and assign the label of the scene with the highest visibility to that point. The results demonstrate that our method can accurately capture the unoccluded parts in each scene, leading to clean background and 360$^\circ$ foreground object rendering. Moreover, it also implies that our simple yet versatile concept of visibility field can accurately quantify visibility information in 3D space, which may benefit future research in various fields.

\subsection{Visibility field smoothing}
\label{sec:visibility-smoothing}
\begin{figure*}
    \centering
    \includegraphics[width=\linewidth]{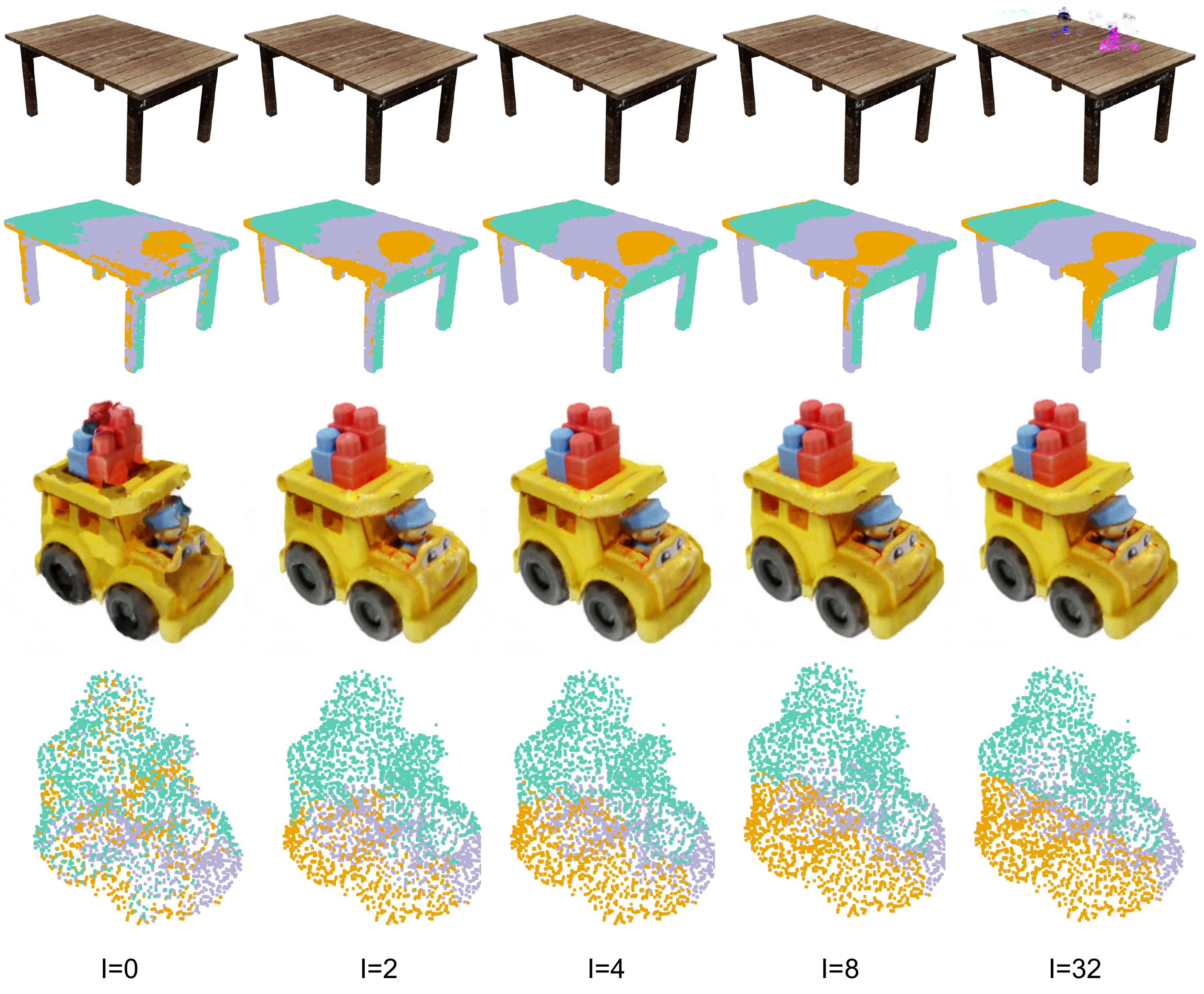}
    \caption{\textbf{Ablation on number of iteration for visibility field smoothing.} Also refers to Fig.~\ref{fig:ablation_labels} for the details of labels. }
    \label{fig:ablation_smoothing}
\end{figure*}
In the formulation of visibility (Eq.~(\ref{eq:visibility_field_sum}) in the main paper), $V_l(\mathbf{x})$ exhibits a step function near the object surface (\eg, from visible to invisible), resulting in discontinuous changes in visibility near the surface. Such discontinuous values make the visibility of the object surface ambiguous between different scenes, leading to an ineffective comparison of visibility between scenes. To address the issue of discontinuity, we smooth the visibility field in our implementation. Specifically, the visibility field is implemented by 3D grid, we apply a discrete Laplacian smoothing to its grid points. For the visibility $v_{d,h,w}$ on the grid point, at each iteration its is updated as:
\begin{equation}
    v_{d,h,w} = \frac{1}{6} \sum v_{\mathcal{N}(d, h, w)} ,
\end{equation}
where $\mathcal{N}(d, h, w)$ denotes the six adjacent grid points to the grid $(d, h, w)$, \ie, $(d\pm1, h, w), (d, h\pm1, w), (d, h, w\pm1)$. We repeat the above update $I$ times to obtain the smoothed visibility field for subsequent fusion computation. 

We present ablation study on the smoothing iterations $I$ of the visibility field in Fig.~\ref{fig:ablation_smoothing}. The visibility field is implemented as a $64^3$ grid.
We can observe that without smoothing, the scene labels near the object surface are very noisy, which also indicates that the visibility values near the surface are ambiguous without smoothing, resulting in the incorrect selection of scenes with higher visibility. As we increase the number of iterations for smoothing, we can observe that the scene labels and rendering results become smoother. This suggests that it pays more attention on the overall visibility of the surrounding area rather than the visibility of just a single point. However, excessive smoothing (\ie, smoothing 32 times for a $64^3$ grid) can make the visibility between scenes too similar to distinguish, resulting in artifacts. Specifically, our setup assumes that the main difference between scenes lies in the foreground, while the background remains consistent. Smoothing can be seen as averaging with the surrounding areas, hence excessive smoothing can weaken the differences between foreground objects, causing the calculated visibility field to become similar across all scenes and affecting the scene fusion result.

In practice, we perform $I=4$ iterations of smoothing for all experiments.

\bibliography{main}
\bibliographystyle{main}

\end{document}